\documentclass{bmvc2k}
\usepackage{multirow}
\usepackage{booktabs} 
\usepackage{amssymb}
\usepackage{diagbox}

\title{Not All Words are Equal: Video-specific Information Loss for Video Captioning}

\addauthor{Jiarong Dong}{dongjiarong@ict.ac.cn}{12}
\addauthor{Ke Gao}{kegao@ict.ac.cn}{1}
\addauthor{Xiaokai Chen}{chenxiaokai@ict.ac.cn}{12}
\addauthor{Junbo Guo}{guojunbo@ict.ac.cn}{1}
\addauthor{Juan Cao}{caojuan@ict.ac.cn}{1}
\addauthor{Yongdong Zhang}{zhyd@ict.ac.cn}{1}

\addinstitution{
Institute of Computing Technology,\\
Chinese Academy of Sciences,\\
Beijing, China
}

\addinstitution{
University of Chinese Academy\\
of Sciences,\\
China
}

\runninghead{Dong, Gao \MakeLowercase{\etal}}{video-specific information loss for video captioning}

\def\etal{\emph{et al}\bmvaOneDot}

\begin{document}

\maketitle

\begin{abstract}
An ideal description for a given video should fix its gaze on salient and representative content, which is capable of distinguishing this video from others. However, the distribution of different words is unbalanced in video captioning datasets, where distinctive words for describing video-specific salient objects are far less than common words such as 'a' 'the' and 'person'. The dataset bias often results in recognition error or detail deficiency of salient but unusual objects. To address this issue, we propose a novel learning strategy called Information Loss, which focuses on the relationship between the video-specific visual content and corresponding representative words. Moreover, a framework with hierarchical visual representations and an optimized hierarchical attention mechanism is established to capture the most salient spatial-temporal visual information, which fully exploits the potential strength of the proposed learning strategy. Extensive experiments demonstrate that the ingenious guidance strategy together with the optimized architecture outperforms state-of-the-art video captioning methods on MSVD with CIDEr score $87.5$, and achieves superior CIDEr score $47.7$ on MSR-VTT. We also show that our Information Loss is generic which improves various models by significant margins.
\end{abstract}

\section{Introduction}
\label{sec:intro}
Video captioning aims at generating both semantically and syntactically correct descriptions for a video. An ideal description for a given video should fix its gaze on salient and representative content, which is capable of distinguishing this video from others. Inspired by the recent progress of machine translation, LSTM network combined with attention mechanism has dominated this task~\cite{yao2015describing,song2017hierarchical}. In this case, the concept recognition is accomplished during the sentence generation process. Existing video captioning models are mainly optimized to maximize the probability of every ground truth word under cross entropy loss learning strategy. However, due to unbalanced distribution of word frequencies, the common words comprise the majority of the training loss and dominate the gradient. Consequently, models rarely achieve satisfactory performance to predict representative and distinctive words, which provide most information of given videos.

\begin{figure}
	\begin{center}
		\includegraphics[width=1.0\linewidth]{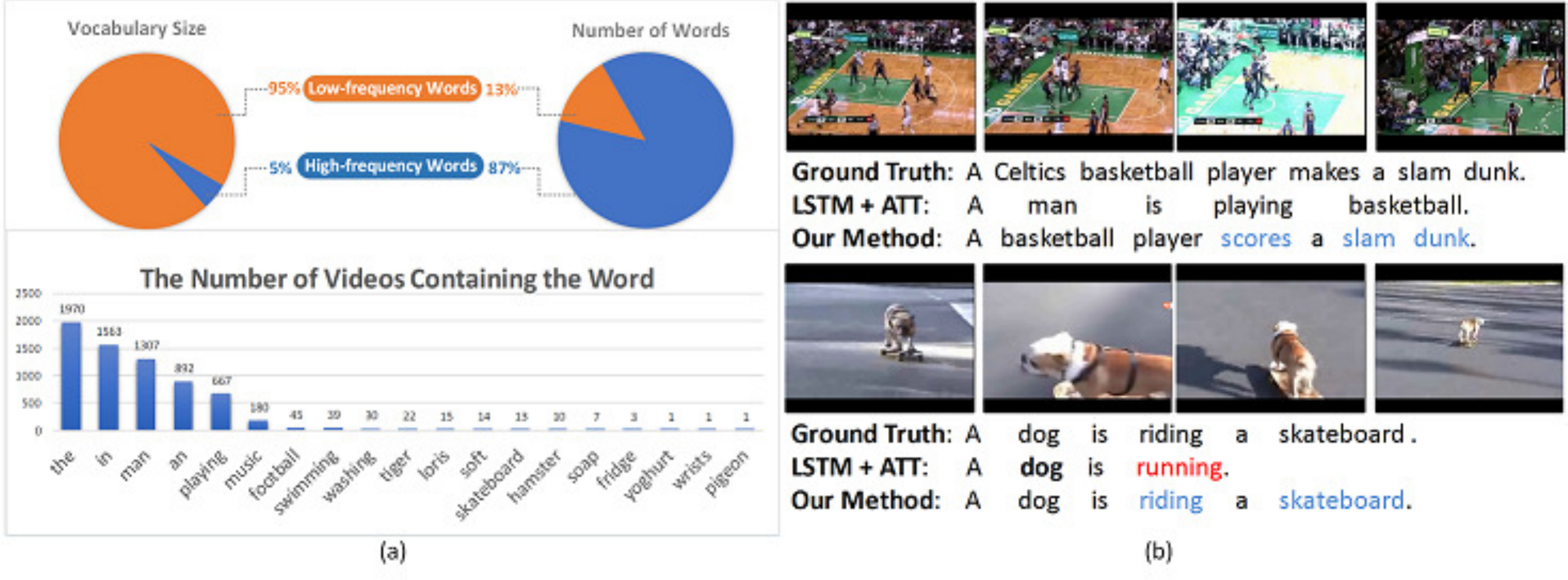}
		\caption{Illustration of description ambiguity in video captioning caused by the bias of dataset corpus. (a) The pie chart is about Distribution of vocabulary size and the number of words of the training corpus. The bar chart below is about the number of videos containing different words. (b) Examples of video description generation.}
		\label{fig:illustration}
	\end{center}
\end{figure}

Figure \ref{fig:illustration}(a) shows a data analysis about MSVD dataset, which is also the case in most video captioning datasets. The top $5\%$ highest frequency words occupy almost $90\%$ content of training corpus. That is, the captioning model spends the vast majority of effort on learning just very small subset of the vocabulary. Moreover, the most of high-frequency words are function words and common semantic words, which present limited information about the video-specific content. For example, the most frequent word "the" appears in all videos of MSVD, whereas the most of distinctive words just appear in a few specific videos. Directly training with this unbalanced data, the existing methods often suffer from description ambiguity, including detail deficency and recognition error~\cite{Zhang2017TaskDrivenDF}. As shown in Figure \ref{fig:illustration}(b), the traditional method(LSTM + ATT for short in this paper) wrongly predicts 'riding a skateboard' as 'running'.     

To address the above issues, we propose a novel learning strategy called Information Loss to focus on learning the representative and distinctive words. Specially, we calculate importance value of every word for the content of given videos, which takes into account both information relevance and information content. Information relevance measures the degree of relevance between a word and the gist of the video. Information content corresponds to the discriminate degree of a word. Therefore, the distinctive words about the salient visual content would obtain high importance value, whereas common words and irrelevant words have low value. Then, the loss of every word is dynamically resized based on this importance value. In this case, the model pays more attention to learn video-specific informative words in every video sample, while the commonly used words can be well learned in overall captioning corpus.

In order to further exploit the potential strength of Information Loss, we established a new framework with hierarchical visual features and an optimized hierarchical attention mechanism. The optimized hierarchical attention model is established to adaptively extract the most important visual features from inter-and-intra video frames, which is the foundation of generating informative video descriptions.

In summary, the main contributions of this paper are: 1) A novel learning strategy called Information Loss is proposed to alleviate the description ambiguity problem caused by the bias of captioning corpus. 2) Moreover, an optimized framework consists of hierarchical visual representations and corresponding hierarchical attention is established to fully exploit the potential strength of the proposed learning strategy. 3) Extensive experiments demonstrate that the optimized structure combined with ingenious guidance can outperform state-of-the-art video captioning methods. 

\section{Related Work}
\noindent\textbf{Model structure for video captioning }Most existing methods can be divided into the bottom-up methods~\cite{huang2012multi,ThomasonVGSM14} and the top-down methods~\cite{venugopalan2014translating,jin2016describing}. The mechanism of the bottom-up method is intuitive. Some predefined concepts are recognized only on visual signals, and then linked into a sentence with fixed language templates. Captions generated in this way often lack flexibility. The top-down methods are inspired by the recent progress of machine translation, where LSTM network combined with attention mechanism has dominated video captioning. The attention mechanism is exploited based on the observation that not all parts of a video are equally important, which has been successfully applied in video captioning to select important frame-level features~\cite{yao2015describing,song2017hierarchical}. There are few works focusing on both object-level and frame-level features in video captioning. Tu \textit{et al.}~\cite{xishan} explore spatial and temporal attention. However, they ignore the interaction between objects over time. Therefore, we establishes the connection of objects between frames to focus on the informative and discriminative visual information of the entire video.         

\begin{figure*}
	\begin{center}
		\includegraphics[width=1.0\textwidth]{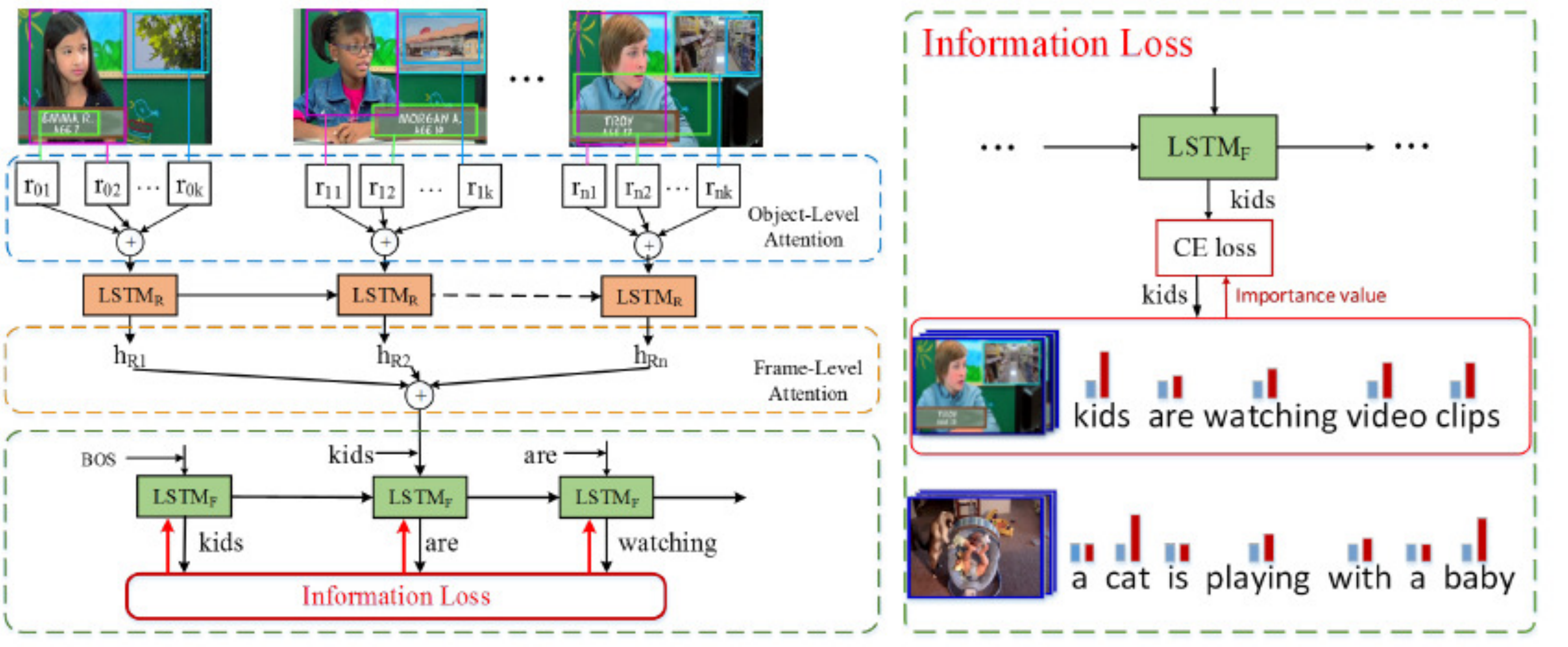}
	\end{center}
	\caption{The framework of our Hierarchical Attention Model with Information Loss(HA\_IL). The hierarchical attention mechnism established on hierarchical visual features captures the most salient visual information. Compared to traditional learning strategy(blue bar), the Information Loss focus on learning the relationship between 'kids' and video content with high importance value(red bar) of 'kids'.}
	\label{fig:framework}
\end{figure*}

\noindent\textbf{Learning strategy for video captioning }Most of traditional video captioning systems are  trained with the cross-entropy loss~\cite{venugopalan2015sequence,yao2015describing}. Although effective for sequence generation tasks, the cross-entropy loss cannot guarantee the semantic correctness ~\cite{jointly} and informativeness of generated captions. Pan \textit{et al.}~\cite{jointly} proposes relevance loss together with cross entropy loss to enforce the relationship between sentence semantics and visual content. The Reinforcement Learning has been introduced in image captioning~\cite{siqiliu,rennie2016self} to improve models by sentence-level rewards, which needs well pretrained model to initialization. However, the both methods suffer from the imbalance learning problem caused by the dataset bias and have not taken distinctiveness into consideration. Our Information Loss focuses on learning representative and distinctive words to overcome dataset bias, which are complementary for these methods.

\section{Approach}

\subsection{Overall Framework}
As shown in Figure \ref{fig:framework}, the overall framework consists of three important components: hierarchical visual features extraction, hierarchical attention mechanism and Information Loss guided learning strategy. Hierarchical attention mechnism is established on hierarchical visual features to capture the salient spatial-temporal visual content. Information Loss is the objective of our model which aims to address the description ambiguity problem. Details are given in section 3.2, section 3.3, section 3.4 respectively.

\subsection{Hierarchical Visual Features Extraction}

The exploitation of object-level features is a key component of our method to generate informative captions. We use Faster R-CNN~\cite{ren2015faster} to generate a set of semantic region features from every frame. The Faster R-CNN model is pretrained on Visual Genome data~\cite{vendrov2015order} by~\cite{anderson2017bottom} and can recognizes 1600 objects, which produces rich and diverse information. For every video frame, we select the top k most confident objects detected by Faster R-CNN model and extract features of these objects from the pool5 layer. Therefore, we get a set of semantic object-level features of a video, denoted as $VR=\{vr_1,\ldots,vr_n\}$, $vr_i=\{vr_{i1},\ldots,vr_{ik}\},vr_{ij}\in \mathbb{R}^{D1}$, where $vr_{ij}$ means the $j_{th}$ region feature of $i_{th}$ frame. Then we apply ResNet~\cite{he2016deep} to extract global frame-level features, denoted as $VF=\{vf_1,\ldots,vf_n\},vf_i\in \mathbb{R}^{D2}$, where $n$ is the number of frames sampled from each video. In addition, for every video, we apply C3D~\cite{c3d} to extract several clip-level features, denoted as $VC=\{vc_1,\ldots,vc_n\},vc_i\in \mathbb{R}^{D3}$. Finally, a given video is encoded as a feature set $V=\{VR,VF,VC\}$.
\begin{figure}[t]
	\begin{center}
		\includegraphics[width=0.9\linewidth]{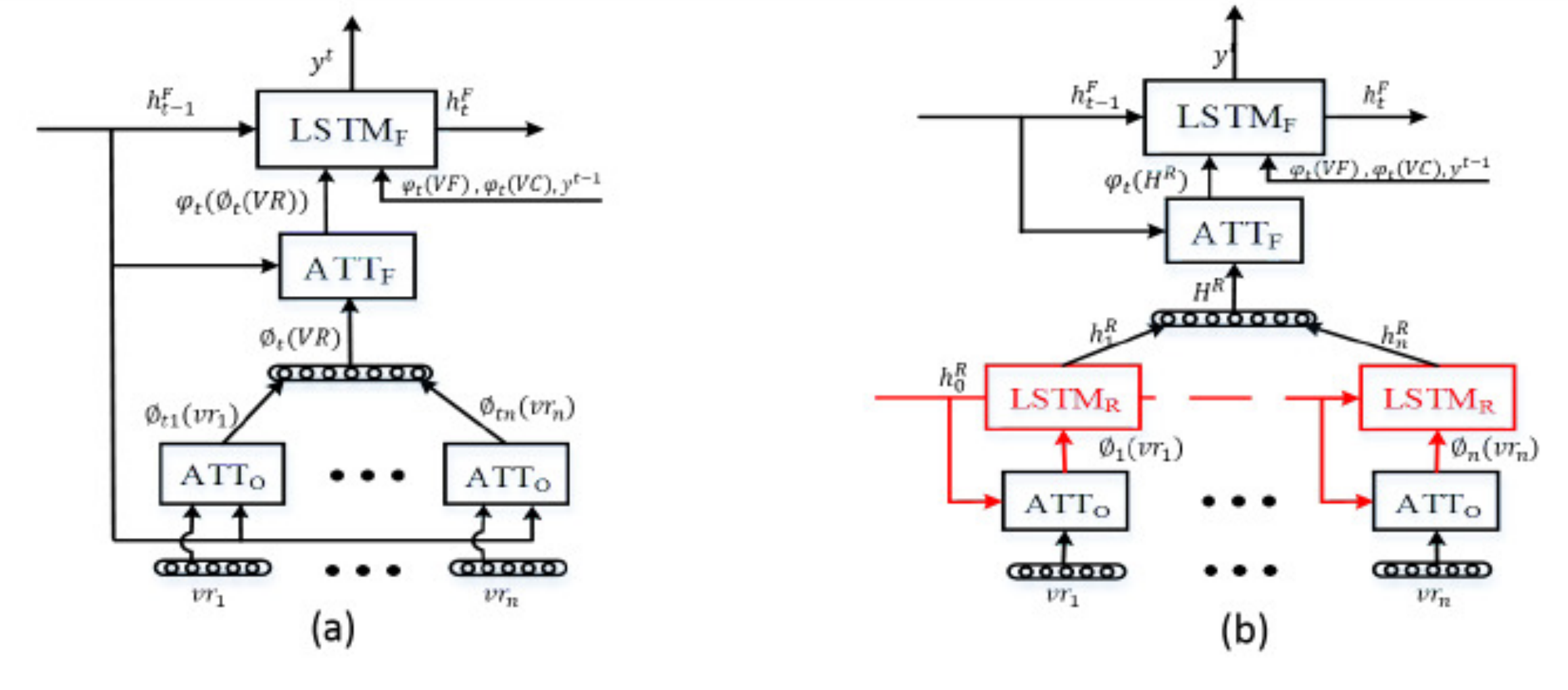}
		\caption{An illustration of hierarchical attention model of STAT(a) and our visual model(b).}
		\label{fig:comparision}
	\end{center}
\end{figure}
\subsection{Hierarchical Attention Mechanism}
Hierarchical attention mechanism is an effective achitecture to exploit hierarchical visual features. However, it is not easy to choose the real salient objects among hundreds of objects for the given video. As shown in Figure \ref{fig:comparision}, different from the previous work~\cite{xishan} which directly attends to objects mostly based on syntactic context, we introduce an extra LSTM to establish the connection of objects between frames and focus on the salient objects based on the video content. Our model is composed of two LSTM layers associated with two attention layers. The first one is denoted as $LSTM_R$ which models temporal relationship of objects in a given video. The second one is denoted as $LSTM_F$ which acts as a language model to establish a syntactic context and generate natural language sentences. The two attention layers are focusing on object-level and frame-level information respectively.

\textbf{Object-level attention with $LSTM_R$ }
The object-level attention with $LSTM_R$ is used to select salient semantic regions. Given the previous hidden state and object-level features at time step $i$, $LSTM_R$ encodes video information until $i_{th}$ frame into the hidden state $h_i^R$:
\begin{equation}
h_i^R = LSTM_R(\phi_i(vr_i),h_{i-1}^R)
\end{equation}
where the visual feature presentation of the $i_{th}$ frame $\phi_i(vr_i)$ is the weighted sum of all the $k$ semantic regions:
$\phi_i(vr_i )=\sum_{j=1}^{k}\alpha_{ij}vr_{ij}$.
The salient objects are emphasized by the attention weight $\alpha_{ij}$ which is obtained from the object-level attention layer:
\begin{equation}
\alpha_{ij}=\frac{\exp(e_{ij})}{\sum_{j=1}^{k}\exp(e_{ij})}\quad \text{with}\quad
e_{ij}=w_e^T \tanh(W_e h_{i-1}^R+U_e vr_{ij}+z_e)
\label{eq:atto}
\end{equation}
where $w_e,W_e,U_e, z_e$ are parameters to be learnt. It is observed from equation (\ref{eq:atto}) that the attention weight of each object is not only determined by their own features but also by the video information of previous frames. Our motivation lying behind this is that the salient objects of a given video are the ones which have important relations to video context.

\textbf{Frame-level attention with $LSTM_F$ }
The frame-level attention is proposed to allow $LSTM_F$ selecting relevant subset of frames while generating words step by step. For $LSTM_F$, the input are the results of frame-level attention $\varphi_t{(H^R)}$, $\varphi_t(VF)$, $\varphi_t(VC)$, the previous hidden state $h_{t-1}^F$, and the previous word $y_{t-1}$:
\begin{equation}
h_t^F=LSTM_F(W_E y_{t-1}, \varphi_t{(H^R)}, \varphi_t(VF), \varphi_t(VC), h_{t-1}^F)
\end{equation}
$h_t^F$ absorbs information from visual content and syntactic context and is used to predict word. $W_E$ is the word embedding matrix. $H^R$ is expanded as $\{h_1^R,...,h_n^R\}$, where $h_i^R$ is the output of $LSTM_R$. Note that the time step i of $LSTM_R$ and the time step t of $LSTM_F$ are irrelevant which reduces the time for training and testing. $\varphi_t{(H^R)}$, $\varphi_t(VF)$ and $\varphi_t(VC)$ have the same function expression with different parameters. For simplicity, we use notation $X$ to represent the input of $\varphi_t{(\cdot)}$. $\varphi_t{(X)}$ is dynamically computed by the weighted sum of the element $x_i$ :
\begin{equation}
\varphi_t{(X)=\sum_{i=1}^{n}\beta_{ti}x_{i}}\quad \text{with}\quad
\beta_{ti}=\frac{\exp(m_{ti})}{\sum_{i=1}^{n}\exp(m_{ti})}\text{,}\quad m_{ti}=w_m^T \tanh(W_m h_{t-1}^F+U_m x_{i}+z_m)
\end{equation}
where $w_m,W_m,U_m, z_m$ are parameters to be learnt. The attention weight $\beta_{ti}$ reflects the importance of the $i_{th}$ frame features given $h_{t-1}^F$. The predicted word $y_t$ is obtained through a single hidden layer: $y_t=softmax(U_yh^F_t)$

\subsection{Information Loss}
The Information Loss is designed to alleviate description ambiguity problem. For clarification, we first briefly describe the traditional objective of video captioning. Given the previous groud truth word $y_{t-1}$ and visual features $V$, the cross entropy loss(CE) is used to maximize the log-likelihood of the next groud truth word: 
\begin{equation}
L_{C}(s,V) =-\sum_{t=1}^{T} \log p(y_t|(V,y_{t-1}))
\end{equation}
Where $s=\{y_1,...,y_T\}$ is the given caption. The CE loss is equal for all words. However, the common words can be learned well with total training data while the representative and distinctive words just can be learned from specific video samples. Different words have different importance for expressing video content. The learning strategy should focus more on the video-specific informative words about salient visual content. 

Our Information Loss reshapes the standard cross entropy loss to up-weight the representative and distinctive words for every video sample, which encourages the video captioning model to capture the gist of a video and generate informative descriptions. Specifically, we introduce an importance value to measure both the information relevance and information content of every word for a given video. Then, we add the importance value as modulating factor to the cross entropy loss. 

In existing video captioning datasets, each video clip is annotated with scores of descriptions by different workers. Although each video clip has various content, most persons reach consensus about the salient objects and activities. That is, the representative words corresponding to the main content of a video are more likely to appear in multiple ground truth captions, whereas the irrelevant words occur less frequently. Therefore, we regard word frequency among all ground truth captions of a video as the information relevance:
\begin{equation}
R(y_t|V)=\frac{N_{y_t,V}}{N_V}
\end{equation}
where $N_V$ is the number of captions owned by the given video $V$, $N_{y_t, V}$ is number of captions which contain word $y_t$. The information relevance term encourages the captioning model to capture the main content of a video. However, many commonly used words in the whole captions are also been encouraged. Therefore, we introduce an information content term to suppress the weight of high-frequency words and elevate the weight of informative words. 

Information content is deeply discussed in information theory, which is a synonym for the surprise when a signal is received. We follow the way in information retrieval to calculate information content of every word, which is defined as:
\begin{equation}
I(y_t) = \log(\frac{1}{p(y_t)})\quad \text{with}\quad p(y_t) = \frac{|\{V_{y_t}\}|}{|\{V\}|}
\end{equation} 
where $|\{V_{y_t}\}|$ is the number of videos in which word $y_t$ appears, $|\{V\}|$ is the overall number of videos, and $p(y_t)$ is the frequency that word $y_t$ appears in different videos. The smaller the frequency of word $y_t$, the larger quantity of information carried by $y_t$. Combining the information content term and information relevance term induces the importance value of a word: $f(y_t, V) = R(y_t|V)^\gamma I(y_t)$. $\gamma \geq 0$ is a tunable parameter which smoothly adjusts the interreation of this two terms. When $\gamma=0$, the importance value is equal to information content. While when $\gamma$ is larger, the information relevance has more influence on the importance value, and the importance value of low frequency words of a given video will be rapidly diminished. We add the importance value as modulating factor to the cross entropy loss and define the Information Loss as:
\begin{equation}
L_I(s,V) = -\sum_{t=1}^{T}(1 + \lambda f(y_t, V))\log p(w_t|V,y_{t-1})
\end{equation}
$\lambda$ is the trade-off parameter. This loss focus more on important words than common words based on the importance value. Intuitively, the captioning model needs to matter the visual properties and generate more percision words to reduce the information loss. 

\section{Experiments}
\subsection{Dataset and Evaluation Metrics}
We conduct extensive experiments on two widely used video captioning benchmarks: MSVD~\cite{guadarrama2013youtube2text} and MSR-VTT~\cite{xu2016msr}. MSVD contains 1970 Youtube video clips with around 40 human annotated descriptions per clip. Following the standard setting provided by ~\cite{yao2015describing}, we takes 1200 videos for training, 100 videos for validation and the remainder for testing. MSR-VTT contains 10000 video clips with 20 human annotated descriptions per clip. We use the public splits: 6513 for training, 497 for validation and 2990 for testing. To evaluate the performance, we employ three commonly used evaluation metrics, including BLEU~\cite{papineni2002bleu}, METEOR~\cite{denkowski2014meteor}, and CIDEr~\cite{vedantam2015cider}. CIDEr gives higher score for video-specific n-grams than generic n-grams~\cite{vedantam2017context}, which is used to measure the effectiveness of our method. 
\begin{table*}
	\begin{center}
		\begin{tabular}{c|c|c|c|c|c|c}
			\hline
			\multirow{2}*{Models}
			& \multicolumn{3}{|c|}{MSVD} & \multicolumn{3}{|c}{MSR-VTT}\\
			\cline{2-7}
			& BLEU-4 & METEOR & CIDEr & BLEU-4 & METEOR & CIDEr\\
			\hline
			NA(R+C+F)&49.7&33.6&76.1&40.8&27.0&45.2\\
			TA-B1(R+C)&48.5&33.5&74.5&39.9&26.9&44.9 \\
			TA-B2(R+C+F)&50.2&34.6&77.0&41.4&27.3&45.3 \\
			\hline
			HA(R+C+F)&\textbf{52.3}&\textbf{35.4}&\textbf{82.9}&\textbf{42.1}&\textbf{27.4}&\textbf{46.0}\\
			\hline
		\end{tabular}
	\end{center}
	\caption{Performance comparison between baseline models and HA on the MSVD and MSR-VTT datasets. Here, R, C, F are shorted for ResNet, C3D, Faster R-CNN features.}
	\label{tab:baseline}
\end{table*}

\begin{table}
	\parbox{0.55\linewidth}{
		\begin{center}
			\begin{tabular}{c|c|c|c}
				\hline
				Models&BLEU-4&METEOR&CIDEr\\
				\hline
				S2VT&48.8&32.5&69.7\\
				S2VT\_IL&49.1&33.3&73.2\\
				HA&52.3&35.4&82.9\\
				HA\_IL&\textbf{54.1}&\textbf{36.6}&\textbf{87.5}\\
				\hline
			\end{tabular}
			\caption{Performance comparison between different models trained with CE and IL}
			\label{tab:il}
	\end{center}}
	\parbox{0.4\linewidth}
	{\begin{center}
			\begin{tabular}{l|c|c|c}
				\hline
				\diagbox{$\lambda$}{$\gamma$} &1&2&3\\
				\hline
				0.3&83.6&84.4&83.3\\
				0.5&85.3&\textbf{87.5}&84.3\\
				0.7&83.7&85.6&83.2\\
				1.0&83.1&84.6&82.4\\
				\hline
			\end{tabular}
			\caption{The effect of parameter}
			\label{tab:gamma}
	\end{center}}
\end{table}

\subsection{Implementation Details}
We tokenize video captions and preserve the words which appear at least 5 times in training dataset while other words are replaced by word UNK. As result, we obtain a vocabulary size of 3656 for MSVD dataset and 8760 for MSR-VTT dataset. The object-level features is extracted by Faster R-CNN with dimension 2048. We set the number of objects from each frame to be 16. For frame-level features, we adopt the 2048-dimension pool5 feature from ResNet101 and the 4096-dimension fc6 feature from C3D. We uniformly sample 40 frames for each video with interval 8. We use a wider interval for long videos and pad zero frames for short videos. We empirically set the dimension of all the hidden units to 512. Dropout is used in both the input and output of two LSTM layers. We apply ADAM algorithm~\cite{kingma2014adam} with learning rate $10^{-4}$ to optimize our model under both cross entropy loss and Information Loss. We anneal the learning rate by a factor of 0.8 every 30 epochs. $\lambda$ and $\gamma$ are set to be 0.5 and 2 respectively. The sensitivity of $\lambda$ and $\gamma$ will be discussed in Section 4.3.

\begin{table}
	\parbox{0.48\linewidth}
	{\begin{center}
			\begin{tabular}{c|c|c|c}
				\hline
				Models&B@4&M&C\\
				\hline
				TA&41.9&29.6&51.7\\
				MAM-RNN&41.3&32.2&53.9\\
				STAT&51.1&32.7&67.5\\
				LSTM-TSA&52.8&33.5&74.0\\
				hLSTMat&53.0&33.6&73.8\\
				\hline
				HA\_IL&\textbf{54.1}&\textbf{36.6}&\textbf{87.5}\\
				\hline
			\end{tabular}
			\caption{Performance on MSVD}
			\label{tab:msvd}
	\end{center}}
	\hfill
	\parbox{0.48\linewidth}{
		\begin{center}
			\begin{tabular}{c|c|c|c}
				\hline
				Models&B@4&M&C\\
				\hline
				TA&35.2&25.2&-\\
				STAT&37.4&26.6&41.5\\
				SA-LSTM&38.7&26.9&45.9\\
				v2t\_navigator&40.8&28.2&44.8\\
				dense caption&41.4&28.3&48.9\\
				\hline
				HA\_IL&\textbf{41.9}&27.9&47.7\\
				\hline
			\end{tabular}
			\caption{Performance on MSR-VTT}
			\label{tab:msrvtt}
		\end{center}
	}	
\end{table}
\subsection{Experiment Analysis}

\textbf{Evaluation of hierarchical attention model} Table \ref{tab:baseline} compares the performance of Hierarchical Attention Model (HA) and baseline models.
NA applies none attention with mean-pooled features fed to every step of LSTM. TA-B1 applies temporal attention mechnism and uses none object-level features. TA-B2 applies temporal attention with hierarchical features. All of these models are trained under cross entropy loss. Compared to TA-B1, TA-B2 achieves better performance because of object-level features. Compared to NA, TA-B2 achieves very limited improvements which shows that temporal attention cannot exploit the hierarchical visual features effectively. On the contrary, our HA method outperforms NA model by a large margin on all metrics, such as over 8.9\% relative improvement on CIDEr. 

\textbf{Evaluation of information loss} To study the generalization ability of proposed Information Loss (IL) method, we have trained S2VT~\cite{venugopalan2015sequence} model and our HA model under Information Loss(IL) learning strategy, namely, S2VT\_IL and HA\_IL. S2VT is our implemetion of ~\cite{venugopalan2015sequence} under cross entropy loss. Results are shown in Table \ref{tab:il}. Compared to models learned by cross entropy loss, both performance of S2VT\_IL and HA\_IL are improved in all metrics when learned by our Information Loss. Especially, HA\_IL improved by larger margin than S2VT\_IL, which indicates that our HA model provides stronger visual representation to further reach the potentials of Information Loss. Overall, these results confirms the generalization and effectiveness of our Information Loss. 

\textbf{Study on parameters $\gamma$ and $\lambda$} Our Information Loss introduces two new hyperparameter, the tunable parameter $\gamma$ and tradeoff parameter $\lambda$. Results using different $\gamma$ and $\lambda$ are shown in Table \ref{tab:gamma}. When $\lambda=0$, our loss is equivalent to cross entropy loss which is the result of HA model. As $\lambda$ increases, the loss pays more attention in words with high importance value. With $\lambda=0.5$, the IL yields the highest CIDEr score. $\gamma$ is used to adjust the interation of information relevance and information content. When $\gamma=2$, the two terms have harmonious collaboration. When $\gamma=2, \lambda=0.5$, the model achieves the best score. 

\begin{figure*}
	\begin{center}
		\includegraphics[width=1.0\textwidth]{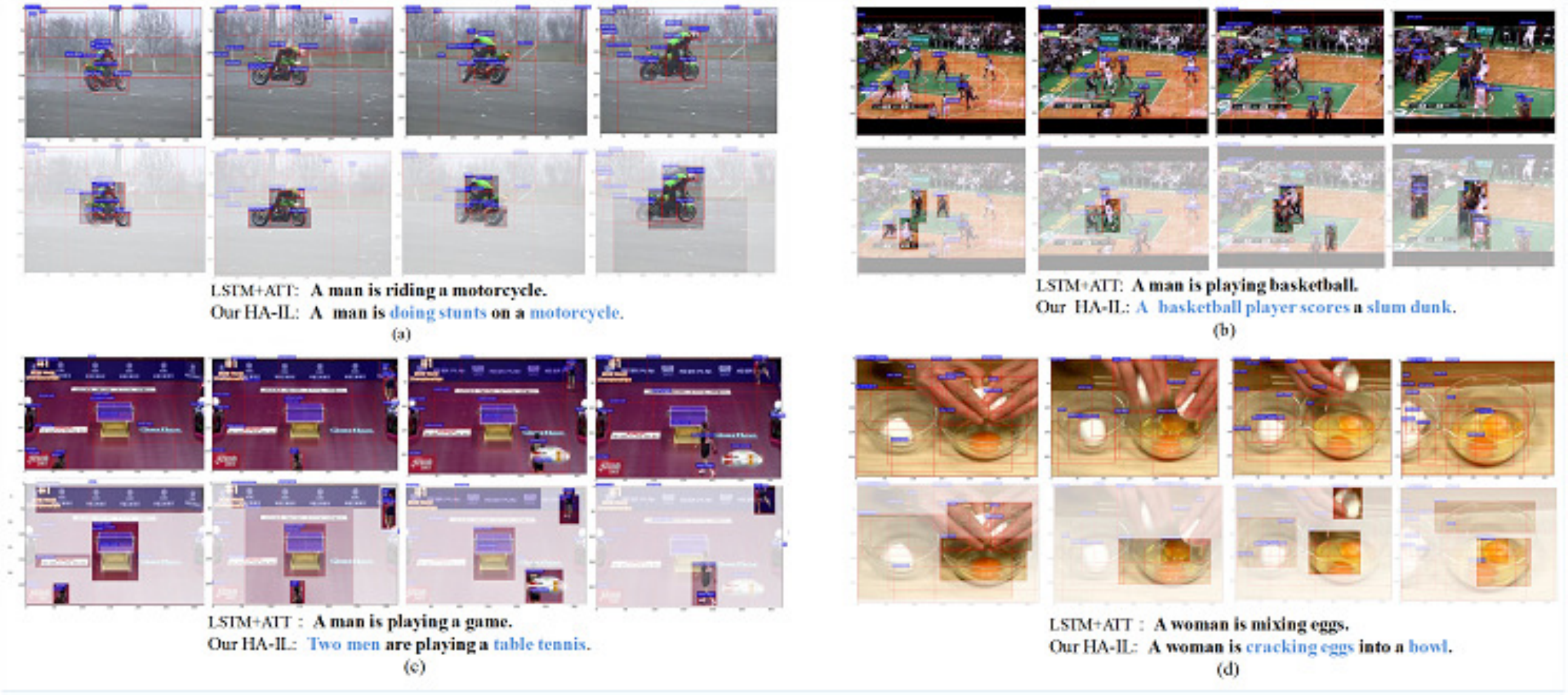}
	\end{center}
	\caption{Visualization of HA\_IL. For each frame, we visualize the attention weights of image regions with the degree of transparency. The less important region is covered with white color more heavily whereas the most salient region remains unchanged. It's clear that our proposed method is capable of focusing on the video-level salient objects.}
	\label{fig:visualization}
\end{figure*}

\textbf{State-of-the-art models }On MSVD dataset, we compare  HA\_IL with five methods: TA~\cite{yao2015describing}, MAM-RNN~\cite{carles}, STAT~\cite{xishan}, LSTM-TSA~\cite{pan2016video}, hLSTMat~\cite{song2017hierarchical}. TA is the first work to employ temporal attention in video captioning. MAM-RNN exploits region level attention. STAT explores object level features with spatial-temporal attention. LSTM-TSA incorporates semantic attributes with RNN frameworks. hLSTMat proposes an adjusted temporal attention. On MSR-VTT dataset, we conduct comparision with five methods : TA, STAT, SA-LSTM~\cite{dong2016early}, v2t\_navigator~\cite{jin2016describing}, dense caption~\cite{shen2017weakly}. SA-LSTM is the baseline method of MSR-VTT. v2t\_navigator is the champion of the MM16 VTT challenge. dense caption focuses on dense video captioning and achieves the best result of MSR-VTT.

\textbf{Overall results} As shown in Table \ref{tab:msvd}, our HA\_IL model outperforms previous methods in all metrics by a large margin on MSVD, such as $18\%$ improvement in CIDEr. 
The performance of our base model HA is better than STAT, suggesting that our hierarchial attention model associated with two LSTM layers is easier to capture the gist of a video. These results confirm the effectiveness of our HA\_IL. As shown in Table \ref{tab:msrvtt}, our HA\_IL model achieves competitive results on MSR-VTT dataset, such as 41.9 on BLEU-4. Notably, v2t\_navigator and dense caption use more kinds of features and execute a lot of engineering efforts such as sentence re-ranking. Hence, we consider that our method achieves comparable performance with less features.  It's worth mentioning that our Information Loss can be used for different base models without any modification. Figure \ref{fig:visualization} visualizes four video examples with detected semantic regions to show how our HA\_IL works. Sentences generated by baseline model and HA\_IL are also provided. It's distinct to see that our model successfully focuses on the video-level salient objects and is able to capture salient details('doing stunts', 'slum dunk', 'cracking'). Especially, although the two men in the example (c) are very small, our visual model attends to them across the entire video. These examples further confirm that our optimized framework and proposed training strategy can jointly capture the gist of the video and generate informative descriptions about given videos.

\section{Conclusion}
This paper observes that the unbalanced distribution of words is a significant cause of recognition error and detail deficiency in video captioning. To address this issue, we propose a novel strategy called HA\_IL. HA\_IL uses Information Loss to alleviate the imbalance learning problem and hierarchical attention mechnism to generate video representation. Our HA\_IL outperforms state-of-the-art video captioning methods with relative improvements $18.2\%$ in CIDEr on MSVD dataset. What' s more , comparative studies show that our Information Loss is generic for various models.

\section*{Acknowledgement}
This work was supported by the National key research and development program (2017YFC0\
820601), National Nature Science
Foundation of China (61571424, U1703261), Beijing science and technology project (Z171100000117010), Beijing Municipal Natural Science Foundation Cooperation Beijing Education Committee: No. KZ 201810005002.

\bibliography{egbib}
\end{document}